\pdfoutput=1

\documentclass[11pt]{article}

\usepackage[final]{acl}
\usepackage{amsmath}
\usepackage{adjustbox}
\usepackage{booktabs}
\usepackage{makecell}
\usepackage{times}
\usepackage{latexsym}
\usepackage{tabularx}
\usepackage{booktabs}
\usepackage[T1]{fontenc}

\usepackage[utf8]{inputenc}
\usepackage{enumitem}
\usepackage{microtype}
\usepackage{multirow}
\usepackage{inconsolata}

\usepackage{graphicx}

\title{Exploring and Evaluating Multimodal Knowledge Reasoning Consistency of Multimodal Large Language Models}

\author{
  \textbf{Boyu Jia\textsuperscript{1}},
  \textbf{Junzhe Zhang\textsuperscript{2}},
  \textbf{Huixuan Zhang\textsuperscript{2}},
  \textbf{Xiaojun Wan\textsuperscript{2}\thanks{Corresponding author. }}
\\
\\
  \textsuperscript{1}School of Software and Microelectronics, Peking University
\\
  \textsuperscript{2}Wangxuan Institute of Computer Technology, Peking University
\\
  \small{
    \texttt{\{jiaboyu, junzhezhang, zhanghuixuan\}@stu.pku.edu.cn, \{wanxiaojun\}@pku.edu.cn} 
  }
}

\begin{document}
\maketitle
\begin{abstract}
In recent years, multimodal large language models (MLLMs) have achieved significant breakthroughs, enhancing understanding across text and vision. However, current MLLMs still face challenges in effectively integrating knowledge across these modalities during multimodal knowledge reasoning, leading to inconsistencies in reasoning outcomes. To systematically explore this issue, we propose four evaluation tasks and construct a new dataset. We conduct a series of experiments on this dataset to analyze and compare the extent of consistency degradation in multimodal knowledge reasoning within MLLMs. Based on the experimental results, we identify factors contributing to the observed degradation in consistency. Our research provides new insights into the challenges of multimodal knowledge reasoning and offers valuable guidance for future efforts aimed at improving MLLMs.
\end{abstract}

\section{Introduction}
\label{sec:Introduction}
Currently, multimodal large language models (MLLMs)\citep{yin2023survey} have garnered significant attention for their ability to integrate multiple data modalities, such as text, images, and audio, thereby enhancing the model's capability in cross-modal understanding and reasoning\citep{nie2024mmrel}. Despite the progress MLLMs have made in specific reasoning tasks such as language understanding and image recognition, significant challenges remain in multimodal knowledge reasoning tasks that involve knowledge fusion across modalities. A major limitation is their insufficient ability to effectively integrate knowledge across different modalities, resulting in inconsistencies in reasoning outcomes, making it difficult for MLLMs to maintain reliable performance in complex reasoning tasks.

To evaluate the reasoning capabilities of MLLMs, researchers have proposed numerous benchmark datasets that assess model performance across various tasks\citep{li2024seed, yu2023mm}. However, many of these benchmarks primarily focus on evaluating the model’s ability to interpret superficial visual information, such as object recognition\citep{wu2024v}, multi-class identification\citep{wang2023makes}, and basic image description\citep{fu2024blink}. While these tasks provide insights into the model's perceptual understanding, they fall short in assessing its capability to perform complex reasoning that requires deep integration of both visual and textual knowledge. As a result, existing evaluation frameworks may not fully capture the true reasoning potential of MLLMs, particularly in scenarios where the model needs to synthesize multimodal knowledge to derive nuanced inferences.

\begin{figure}[tbp]
    \centering
    \includegraphics[width=0.5\textwidth]{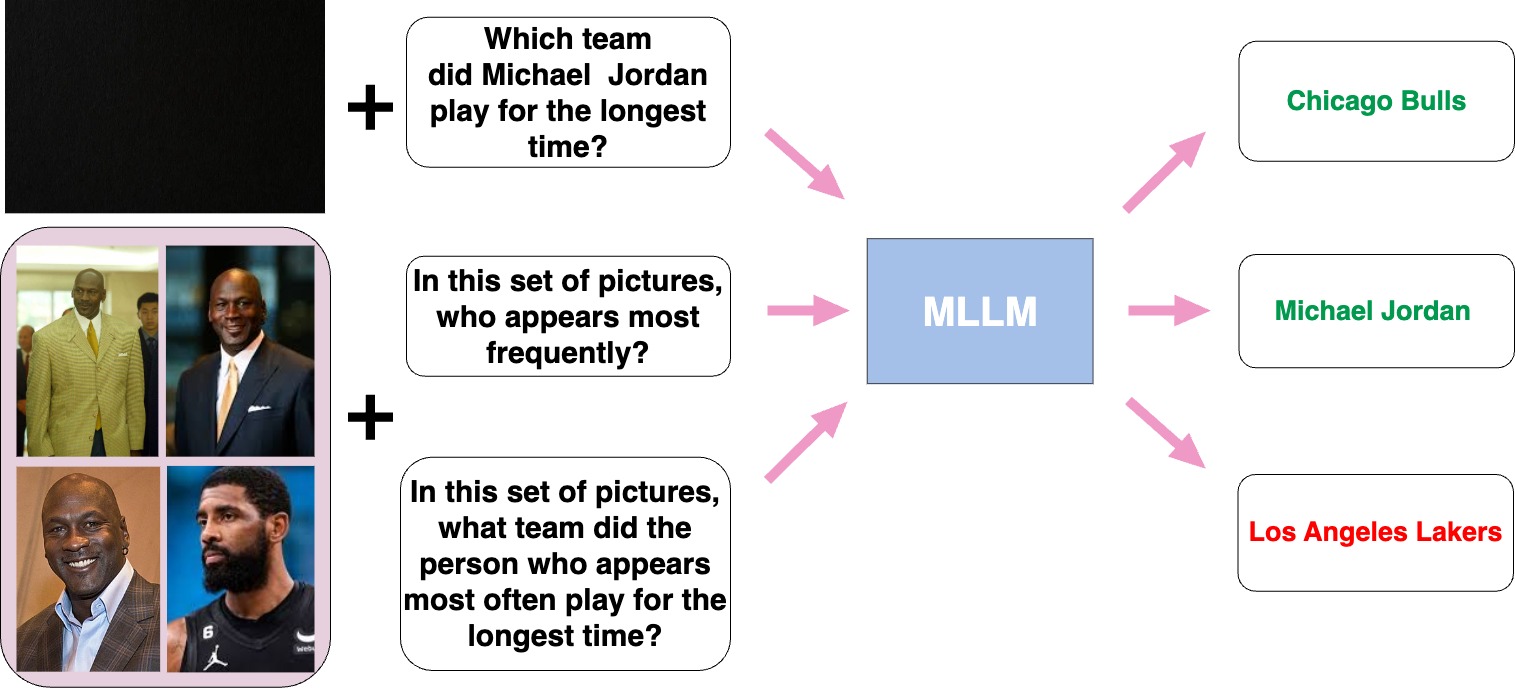}
    \caption{An example of measuring the consistency of a multimodal language model in a multimodal knowledge reasoning task.
    (Given three pictures of Michael Jordan and one picture of basketball star Kyrie Irving, the team Michael Jordan played for the longest time was the Chicago Bulls).}
    \label{fig:mi_example}
\end{figure}

Figure \ref{fig:mi_example} shows an example where model suffer from inconsistency during multimodal knowledge reasoning. When we input a black picture and ask the model about the knowledge chain in the text, the model provides the correct answer. Similarly, when we input three pictures of basketball star Michael Jordan and one picture of basketball star Kyrie Irving, the model successfully completes the visual task of identifying the most frequent character. However, when we combine these two questions to assess the model's ability to reason about multimodal knowledge, the model delivers an incorrect answer. This phenomenon indicates that even when all individual steps in the reasoning chain are correct, the model still struggles to produce a consistent reasoning result, highlighting a failure in maintaining consistency.

Motivated by the above observation, we propose four evaluation tasks (See Section \ref{sec:Task Design} for details of all tasks.) and construct a new dataset to study the consistency problem encountered by MLLM in multimodal reasoning. Specifically, we introduce tasks involving multiple images and multiple reasoning hops to thoroughly investigate this issue. Our dataset can serve as a common benchmark for complex multimodal knowledge reasoning. We systematically evaluate various popular MLLMs using our dataset and analyze the factors contributing to the inconsistency.

The contributions of our work can be summarized as follows: 1) We discover that MLLMs suffer from inconsistency in multimodal knowledge reasoning. 2) We construct a multimodal, multi-image, multi-hop, multi-task dataset for evaluating multimodal knowledge reasoning. \footnote{Our dataset will be released to the community.}. 3) Based on the experimental results, we analyzed the causes of MLLM inconsistency and found that consistency is affected by factors such as the number of inference hops and inference relations.

\section{Related works}
\subsection{Multimodal Large Language Models}
In recent years, the remarkable success of large language models (LLMs) \citep{achiam2023gpt} has significantly influenced the development of multimodal large language models (MLLMs), leading to breakthrough advancements in visual-language alignment. Early works such as CLIP \citep{radford2021learning} and BLIP \citep{li2023blip} established cross-modal pretraining to achieve multimodal ability. Models like Flamingo \citep{alayrac2022flamingo} and BLIP-2 demonstrated strong zero-shot reasoning capabilities by aligning visual features with LLMs. With the advent of models like LLaVA-NeXT \citep{liu2024llavanext}, MiniGPT-4 \citep{zhu2023minigpt}, InstructBLIP \citep{dai2023instructblip}, mPLUG-Owl3 \citep{ye2024mplug}, and Qwen2-VL\citep{wang2024qwen2}, there has been a growing trend of using multimodal instruction fine-tuning data to further enhance the reasoning capabilities of MLLMs in visual-language tasks.

\subsection{Multimodal Large Language Model reasoning}
To evaluate the reasoning capabilities of MLLMs, numerous benchmarks have been introduced. \citet{chen2024m} focuses on visual modality and multihop tasks within single-image scenarios, limiting broader multimodal applicability. \citet{wang2024mementos} includes temporal dimensions in image series reasoning but prioritizes visual tasks over deep multimodal interactions. Similarly, \citet{zhao2024benchmarking} provides a comprehensive multi-image understanding benchmark but lacks complexity for advanced multimodal inference. \citet{li2023fine}, \citet{fu2024blink}, and \citet{xu2024exploring} focus on simple reasoning using image information without addressing complex multimodal tasks. \citet{balesni2024two} investigates inconsistencies between single-hop and multi-hop tasks but only considers textual reasoning. Although some work \citep{wang2024sok} proposes a knowledge benchmark, it lacks an analysis of model performance in knowledge reasoning. Other works, including \citet{chou2024mm} and \citet{zhang2024cross}, explore consistency of model outputs across different modalities but primarily assess alignment between text and vision rather than the deeper integration required multimodal reasoning.

A common limitation is the focus on unidirectional reasoning (e.g., vision-to-text) rather than bidirectional multimodal reasoning. Moreover, there is a lack of systematic analysis on information degradation—a critical issue in multimodal knowledge reasoning where essential details are lost between modalities. Addressing this gap is crucial for enhancing the robustness of MLLMs in real-world applications.

\section{Problem Definition }
\subsection{Multimodal Knowledge Reasoning and Consistency}
\label{sec:Definition}

To clearly define the consistency problem in multimodal knowledge reasoning explored in this study, we adopt a multimodal knowledge definition provided by \citet{zhang2024mc}, where multimodal knowledge is considered a joint representation of visual and textual information. Specifically,  a piece of visual knowledge is denoted as \((i, e)\), where \(i\) represents the image, and \(e\) is the entity recognized from it. (Note that a visual knowledge can also be reversed in image retrieval task, noted as $(e, i)$.) What's more, when there are multiple images to discuss, we use an extended representation $(i_1, ..., i_m, e)$, where $e$ is the recognized entity that appears most often in $\{i_1, ..., i_m\}$. \footnote{This is not a necessary definition but rather a helpful notation in this research.} Similarly, a piece of textual knowledge is expressed as a triple \((s, r, o)\), where \(s\) denotes the subject, \(r\) represents the relation, and \(o\) is the object. 

In our multimodal knowledge reasoning task, to answer a question, multiple pieces of knowledge are concatenated into a chain, namely reasoning chain. For example, to answer a question ``What is the $r$ of the entity in image $i$?", the model needs to first identify the entity in the image, requiring $(i,e)$ knowledge, then get the correct $o$ corresponding $r$ and the entity, requiring $(s,r,o)$ knowledge. The corresponding reasoning chain is shown in Equation \ref{eq:eq2}, where the entity($\bowtie_{e=s}$) concatenates visual knowledge and textual knowledge. 

\begin{equation}
\label{eq:eq2}
(i, e) \bowtie_{e=s} (s, r, o) \Rightarrow (i, r, o)
\end{equation}

Normally, a reasoning chain can be represented as:
\begin{equation}
\label{eq:reasoning}
k_1 \bowtie k_2 \bowtie ... \bowtie k_n \Rightarrow k
\end{equation}
where $k_i$ represents either visual knowledge $(i, e)$($(e, i)$, $(i_1,...,i_m, e)$) or textual knowledge $(s,r,o)$ and concatenated by the same entity ($\bowtie_{e=s}, \bowtie{o=s}$ or $\bowtie_{o=e}$), $k$ is the final knowledge corresponding to a multimodal knowledge reasoning question. 

There are two ways forming a multimodal knowledge reasoning question $q_{k}$ from the above reasoning chain. \textbf{Forward} is giving the beginning and all intermediate relations in $k$ and querying the ending of $k$, while \textbf{Backward} is giving the ending of $k$ and all intermediate relations in $k$ and querying the beginning of $k$. As can be seen, \textbf{Backward} problems are often open with many possible answers, making it more difficult to answer.

Ideally, if a model correctly understands all knowledge $k_i$ in the reasoning chain, it can correctly solve the overall multimodal knowledge reasoning question $q_k$. However, this is not always the case in reality, where models can correctly pass each step in the reasoning chain while still failing to address the overall multimodal reasoning task. We name this phenomenon \textbf{inconsistency}, and the opposite side is \textbf{consistency}, inspired by \citep{zhang2024mc}.

The primary focus of this study is to investigate how well consistency is maintained during multimodal knowledge reasoning. We introduce multiple tasks in Section \ref{sec:Task Design} to thoroughly evaluate consistency. In each experimental task, the following three-step reasoning subtask is performed to evaluate consistency.
\begin{enumerate}
    \item \textbf{Step 1 (Vision Centered Task)}: Asking the model to identify the entity in the image, which focuses on visual knowledge $(i, e)$.
    \item \textbf{Step 2 (Text Centered Task)}: Asking the model to generate the object given subject $s$ and relations $r_1, ..., r_n$, which focuses mainly on textual knowledge reasoning chain.
    \begin{equation}
    \label{eq:textchain}
        \begin{aligned}
        (s_1, r_1, o_1) \bowtie_{o_1=s_2}...\bowtie_{o_{n-1}=s_n}(s_n, r_n, o_n) \\\Rightarrow (s_1,r_1,...,r_n, o_n)
        \end{aligned}
    \end{equation}
    \item \textbf{Step 3 (Multimodal Task)}: Asking the model a question which requires concatenating both visual and textual knowledge.
\end{enumerate}

A model can only be evaluated for consistency using overall multimodal knowledge reasoning question $q_k$ when it correctly understands each component $k_i$ and the textual reasoning chain (\ref{eq:textchain}) in Steps 1 and 2. Otherwise, even if $q_k$ is incorrectly answered, this mistake may simply come from model failing on a certain piece of knowledge $k_i$ or the textual reasoning chain \ref{eq:textchain}, which is nothing surprising. Therefore, to evaluate consistency in multimodal knowledge reasoning, we introduce the \textbf{Consistency Rate (CR)} metric. Let \(S\) be the samples for which all steps $k_i$ and textual reasoning chain (\ref{eq:textchain}) are correctly answered. The CR metric is defined as the proportion of samples in \(S\), for which the overall multimodal knowledge reasoning question $q_k$ also produces the correct answer. The formula is given as follows.

\begin{equation}
\label{eq:eq3}
CR = \frac{|\{ q_k \mid q_k \in S, \text{$q_k$ is correctly answered.} \}|}{|S|}
\end{equation}

It is important to again note that we assess consistency of multimodal knowledge reasoning task only when the model provides correct answers for all steps $k_i$ in the reasoning chain. A failure of multimodal knowledge reasoning under this premise. By utilizing this metric, our study aims to analyze multimodal knowledge reasoning consistency and propose improvements to enhance overall model consistency.

\begin{figure*}[htbp]
    \centering
    \includegraphics[width=0.95\textwidth]{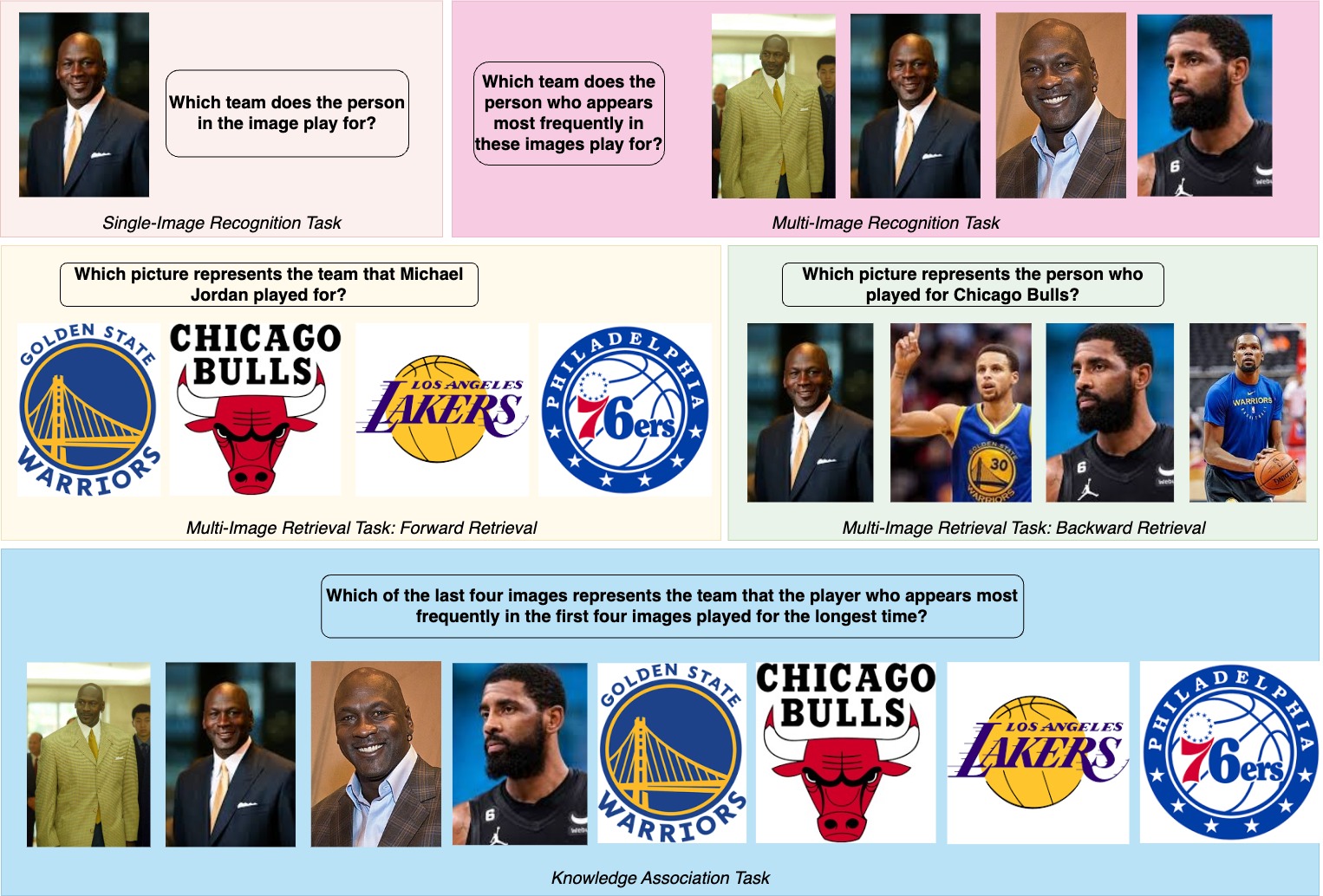}
    \caption{Examples of our multimodal knowledge reasoning tasks.}
    \label{fig:task}
\end{figure*}

\subsection{Task Design}
\label{sec:Task Design}
There are many possible ways of constructing reasoning chain for evaluating consistency. We design four representative tasks for evaluation as follows and present an example of each task in Figure \ref{fig:task}:

\begin{itemize}[leftmargin=*]
 \setlength{\leftmargin}{0pt}
    \item \textbf{Single-Image Recognition:} This task requires first identifying the entity in the image, then performing single or multiple reasoning steps on textual knowledge. The reasoning chain is formulated as:
    \begin{equation}
    \begin{aligned}
        (i, e)\bowtie_{e=s_1}(s_1,r_1,o_1)\bowtie_{o_1=s_2}...\\ 
        \bowtie_{o_{n-1}=s_{n}}(s_n, r_n, o_n) \Rightarrow (i, r_1, ..., r_n, o)
    \end{aligned}
    \end{equation}
    
    \item \textbf{Multi-Image Recognition:} This task is an extended version of single-image one, with multiple images and the model should identify the entity of each image, then select the entity appearing most often in the images and answer corresponding textual knowledge. The reasoning chain is formulated as:
    \begin{equation}
        \begin{aligned}
        (i_1,...,i_m, e)&\bowtie_{e=s_1}(s_1,r_1,o_1) \\\bowtie_{o_1=s_2}... &\bowtie_{o_{n-1}=s_{n}}(s_n, r_n, o_n) \Rightarrow \\&(i_1,...,i_m, r_1, ..., r_n, o)
    \end{aligned}
    \end{equation}
    
    \item \textbf{Multi-Image Retrieval:} The model needs to select the correct image from the given images to answer $q_k$. We consider both forward and backward ways of forming the question. For \textbf{Forward Retrieval}, the task is identifying the correct image representing the object of a textual reasoning chain, and the corresponding reasoning chain is formulated as: 
    \begin{equation}
        \begin{aligned}
            (s_1, &r_1, o_1) \bowtie_{o_1=s_2} ... \bowtie_{o_{n-1}=s_{n}}
            (s_n, r_n, o_n)\\ & \bowtie_{o_n=e} (e, i) \Rightarrow (s_1, r_1,...,r_n, i)
        \end{aligned}
    \end{equation}
    While for \textbf{Backward Retrieval}, the task is identifying the correct image representing the subject of a textual reasoning chain, the corresponding reasoning chain is formulated as:
    \begin{equation}
    \begin{aligned}
        (i, e) &\bowtie_{e=s_{1}} (s_1,r_1,o_1) \bowtie_{o_1=s_2}...\bowtie_{o_{n-1}=s_n} \\
        &(s_n, r_n, o_n) \Rightarrow (i, r_1,...,r_n, o)
    \end{aligned}
    \end{equation}
    Note that $q_k$ is formulated in the backward way in \textit{Backward Retrieval}, so both \textit{Forward Retrieval} and \textit{Backward Retrieval} are image retrieval tasks querying $i$. Please refer to Appendix \ref{sec:A.1} for more details.
    
    \item \textbf{Knowledge Association:} Previous three tasks only involve modality transfer (from textual knowledge to visual knowledge or from visual knowledge to textual knowledge) once, so we would like to evaluate model performance when there are multiple modality transformations. This task combines the \textit{Multi-Image Recognition} task and the \textit{Forward Retrieval} task, requiring the model to associate knowledge by transferring between modalities multiple times. The model needs to correctly identify the images, complete text reasoning, and then complete the \textit{Forward Retrieval} task. The reasoning chain is formulated as follows:
    \begin{equation}
    \begin{aligned}
        (i_1, i_2,...,i_n, e_1) \bowtie_{e_1 = s} (s, r, o)\bowtie_{o=e_2} \\ 
        (e_2, i) \Rightarrow (i_1, ..., i_n, r, i)
    \end{aligned}
    \end{equation}
    
    This task simulates the complex reasoning requirements in real-world scenarios. An example can be found in Figure \ref{fig:task} and more details can be found in Appendix \ref{sec:A.2}.
\end{itemize}


\section{Dataset Construction}
The text data used in the experiments is sourced from the MQuake dataset\citep{zhong2023mquake}, which is designed for knowledge graph editing and contains multiple data instances based on triples $(s, r, o)$, where $s$ represents the subject, $r$ represents the relation, and $o$ represents the object. We construct our data based on the knowledge triples before knowledge editing. The dataset's triple relations cover various levels of reasoning tasks, including two-hop, three-hop, and four-hop reasoning tasks. For each subject $s$ and object $o$ in the triple, we crawled ten relevant images from Google. These images together with the text data triplets constitute the basic dataset $D$ with a size of 3,770. We also decompose all multi-hop data in the original dataset $D$ into two-hop dataset $D_{T}$, with a size of 3,240.  We introduce how the dataset of the four tasks are constructed separately as follows.

\paragraph{Single-Image Recognition}
For \textit{Single-Image Recognition} task, we utilize all four-hop data points in $D$ and construct $n-hop$ reasoning data by truncating first $n$ hops in the four hop question and selecting an image corresponding to the $s$ in the first hop as the image input.

\paragraph{Multi-Image Recognition}
For the \textit{Multi-Image Recognition} task, we utilize all data in two-hop dataset $D_{T}$ but focus only on the first hop to reduce the difficulty of multi-image reasoning. We used GPT-4o to rank the relevance of the crawled images and entities, selecting the top three as relevant images and one irrelevant as interfering item. These four images form the input images.

\paragraph{Multi-Image Retrieval}
For the \textit{Multi-Image Retrieval} task, we also utilize all data in two-hop dataset $D_{T}$. We construct both single-hop (using the first hop) and two-hop questions based on $D_{T}$. We select the image most relevant to the entity to be retrieved as the input and randomly selected other images of the same type of entity as interference options. For text problems, we used GPT-4o to generate two prompts for each type of retrieval reasoning data based on \textit{Forward Retrieval} and \textit{Backward Retrieval} reasoning chains. 

\paragraph{Knowledge Association}
For the \textit{Knowledge Association} task, we reused the images and texts from the \textit{Multi-Image Recognition} and \textit{Multi-Image Forward Retrieval} tasks. Using GPT-4o, we generated two question prompts, requiring the model to complete the \textit{Multi-Image Recognition} task from the first four images and the \textit{Multi-Image Forward Retrieval} task from the second four images.

The amount of data, number of images, and number of reasoning hops for different tasks are shown in the Table \ref{tab:data}.

\begin{table}[htbp]
\centering
\resizebox{0.5\textwidth}{!}{
\renewcommand{\arraystretch}{1.2}
\begin{tabular}{lccc}
\toprule
Reasoning Task & Hops & Data & Images \\
\midrule
\multirow{4}*{\makecell[c]{Single-Image Recognition}} 
    & 1 & 729  & 1 \\
    & 2 & 729  & 1 \\
    & 3 & 729  & 1 \\
    & 4 & 729  & 1 \\
\hline
Multi-Image Recognition & 1 & 3240 & 4 \\
\hline
\multirow{2}{*}{Multi-image Retrieval} & 1 & 3240 & 4 \\
~ & 2 & 3240 & 4 \\
\hline
Knowledge Association  & 1 & 3240 & 8 \\
\bottomrule
\end{tabular}
}
\caption{Data information of different tasks (including Number of Reasoning Hops, Number of Data, and Number of Input Images)}
\label{tab:data}
\end{table}

\begin{table*}[htbp]
\centering
\resizebox{\textwidth}{!}{
\renewcommand{\arraystretch}{1.2}
\begin{tabular}{lccccccccccc}
\toprule
Reasoning Task       & Type      & {LLava-NeXT} & {mPLUG-Owl3} & {GPT-4o} & {Qwen2-VL} & {InstructBLIP} \\
              
\midrule
\multirow{4}*{\makecell[c]{Single-Image\\Recognition}}& single-hop & \underline{74.63}       & 72.45           & \textbf{86.38}            & 74.40       &  31.58      \\
              & two-hop    & \underline{62.05}            & 33.33       & \textbf{83.49}           & 53.74    & 31.58  \\
              & three-hop  & \underline{59.72}        & 27.59      & \textbf{81.01}     & 53.17       &   33.33     \\
              & four-hop   & \underline{60.00}     & 21.15     & \textbf{79.06}    & 49.21   &  25.00 \\
\midrule
Multi-Image Recognition    &       & \underline{76.46}       & 60.41    & \textbf{94.52}      &  /          &     /    \\
\midrule 

\multirow{2}*{\makecell[c]{Multi-Image\\Retrieval(Forward)}}  & single-hop & 21.13        & \underline{85.43}      & \textbf{87.18}        &    /   &/              \\
              & two-hop    & 13.21       & \underline{72.05}     & \textbf{77.69}       & /  &      /         \\
\multirow{2}*{\makecell[c]{Multi-Image\\Retrieval(Backward) }}  & single-hop & 13.57        & \underline{81.12}      & \textbf{82.20}        &  /       &/              \\
              & two-hop    & 10.37  & \underline{71.93}     & \textbf{72.65}       & /  &    /             \\
\midrule
\multirow{1}*{\makecell[c]{Knowledge Association}} & &15.31 &\underline{24.87} & \textbf{70.58} &/ &/\\
\bottomrule
\end{tabular}
}
\caption{Comparison of the consistency performance of different models on different tasks. We label the best result of each task in \textbf{bold} and the second best result with \underline{underline}. / refers to models with no multi-image ability and cannot be evaluated.}
\label{tab:result}
\end{table*}

For the textual questions, the MQuake dataset provides reasoning questions. We also used GPT-4o to generate two distinct questions per data point. To increase diversity and enhance robustness, we randomly selected one question  during testing, allowing us to build a diverse dataset covering multi-hop, multi-image, multi-task knowledge reasoning, for robustly evaluating multimodal knowledge reasoning ability of MLLMs. 

When checking the correctness of an answer, we use aliases to match model output more accurately. More details can be found in Appendix \ref{sec:app-alias}.

\section{Experiments}

\subsection{Experiment Setup}
We selected LLava-NeXT\citep{liu2024llavanext}, InstructBLIP\citep{dai2023instructblip}, Qwen2-VL\citep{wang2024qwen2}, mPLUG-Owl3\citep{ye2024mplug}, and GPT-4o \citep{achiam2023gpt}models to test their consistency capabilities on single-image tasks. For reasoning tasks that require multiple images, we selected LLava-NeXT, mPLUG-Owl3, and GPT-4o models for testing.

\begin{table*}[htbp]
\centering
\resizebox{\textwidth}{!}{
\renewcommand{\arraystretch}{1.2}
\begin{tabular}{lcccccccccc}
\toprule
Reasoning Task       & Type & \multicolumn{4}{c}{\textbf{Stepwise}} \\
\cmidrule(lr){3-7}
                     &      & LLava-NeXT & mPLUG-Owl3 & GPT-4o & Qwen2-VL & InstructBLIP \\
\midrule
\multirow{4}*{\makecell[c]{Single-Image\\Recognition}}   
& single-hop  & 75.57 (+0.94) & 77.55 (+5.10) & 88.10 (+1.72) & 79.00 (+4.60) & 31.58 (+0.00) \\
& two-hop     & 61.45 (-0.60) & 76.19 (+42.86) & 85.32 (+1.83) & 61.92 (+8.18) & 31.58 (+0.00) \\
& three-hop   & 62.50 (+2.78) & 58.62 (+31.03) & 80.78 (-0.23) & 57.07 (+3.90) & 33.33 (+0.00) \\
& four-hop    & 60.00 (+0.00) & 55.77 (+34.62) & 78.82 (-0.24) & 56.02 (+6.81) & 33.33 (+8.05) \\
\midrule
Multi-Image Recognition  
&         & 77.54 (+1.08) & 67.83 (+7.42) & 94.25 (-0.27) & / & / \\
\midrule
\multirow{2}*{\makecell[c]{Multi-Image\\Retrieval(Forward)}}    
& single-hop  & 11.33 (-2.24) & 80.35 (-0.77) & 83.32 (+1.12) & / & / \\
& two-hop     & 9.88 (-0.49) & 70.28 (-1.65) & 72.97 (+0.32) & / & / \\
\midrule
\multirow{2}*{\makecell[c]{Multi-Image\\Retrieval(Backward)}}  
& single-hop  & 19.98 (-1.15) & 84.78 (-0.65) & 87.31 (+0.13) & / & / \\
& two-hop     & 14.02 (+0.81) & 71.22 (-0.83) & 78.58 (+0.89) & / & / \\
\midrule
\multirow{1}*{\makecell[c]{Knowledge Association}}    
&   & 15.39 (+0.08) & 27.05 (+2.18) & 70.13 (-0.45) & / & / \\
\bottomrule
\end{tabular}
}
\caption{The performance on different reasoning tasks using Stepwise prompts. Values in bracelets refer is compared with end-to-end prompts.}
\label{tab:stepwise}
\end{table*}

\subsection{Experiment Results}

The experiment results are presented in Table \ref{tab:result}. As observed, GPT-4o performs best among all models in various tasks. However, its consistency is still worrying in more challenging tasks such as \textit{Multi-Image Retrieval}, indicating that there is still much room for improvement in its multimodal knowledge reasoning consistency.

What's more, other models show even weaker consistency. Although certain models excel on specific tasks, their performance deteriorates largely on others. For instance, LLaVA-NeXT performs competitively in multi-hop reasoning subtask in Single-Image Recognition, achieving strong results compared to other open-source models. However, in the \textit{Multi-Image Retrieval} task, it shows a clear drop in performance, struggling to maintain consistency. Similarly, the InstructBLIP model exhibits overall mediocre performance, and even struggles to achieve favorable results in simpler tasks such as \textit{Single-Image} single-hop reasoning, highlighting its limitations in consistency.

All models perform poorly on \textit{Knowledge Association} task, indicating that multiple transfers between modalities pose significant challenges for even the most powerful MLLMs, underscoring the difficulties of our designed tasks.

\subsection{Analysis}
In this section, we would like analyze the impact of knowledge reasoning hops on consistency, different reasoning relation types, different task types and reasoning process.

\subsubsection{Impact of Knowledge Reasoning Hops}
 We investigate the changes in multimodal knowledge reasoning consistency across different reasoning hops. As is shown in the  \textit{Single-Image Recognition} task of Table~\ref{tab:result}, as the number of hops increases, the models' reasoning consistency gradually declines. This phenomenon suggests that current models lack sufficient capabilities for inferring extended reasoning chains in multi-hop reasoning tasks, leading to cumulative information loss and a failure to maintain consistency throughout the inference process.

\subsubsection{Impact of Relation Types}
\begin{figure}[tbp]
    \centering
    \includegraphics[width=0.5\textwidth]{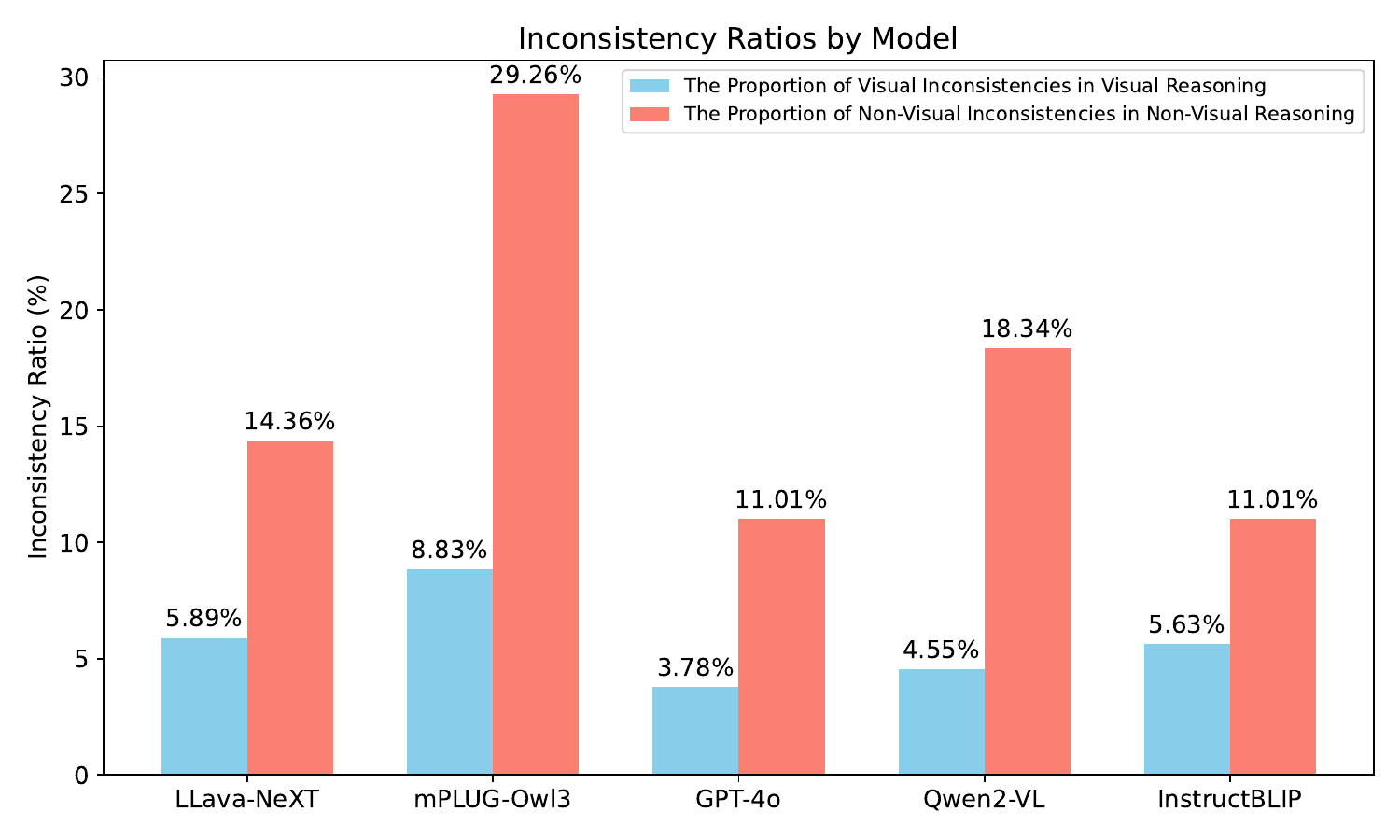}
    \caption{Inconsistency rate of different relation types in different models}
    \label{fig:inconsistency}
\end{figure}

We would like to investigate whether different types relations $r$ affect consistency. Specifically, we classify different relations into two types: relations with clear visual associations (e.g., "nationality" and "genre") and relations with no clear visual associations (e.g., "author" and "creator"). We calculate the inconsistency rate on \textit{Single-hop Single-Image Recognition} task within each relation type and present the result in Figure \ref{fig:inconsistency}.

Our results indicate that relations with clear visual associations exhibit higher consistency rates, while relations that rely on non-visual cues and do require external knowledge often exhibit lower consistency rates. We attribute this to the presence of clear visual cues, which establish direct and stable mappings between visual inputs and corresponding attributes. For example, if an image shows a person wearing a soccer jersey, models may correctly answer the job of this person more easily. Models may achieve accurate outputs without complex intermediate reasoning through utilizing image information when faced with a relation with clear visual associations. Detailed statistics are presented in Appendix \ref{sec:B.1}.

\begin{table*}[htbp]
\centering
\resizebox{\textwidth}{!}{
\renewcommand{\arraystretch}{1.2}
\begin{tabular}{lcccccccccc}
\toprule
Reasoning Task       & Type & \multicolumn{4}{c}{\textbf{VE}} \\
\cmidrule(lr){3-7}
                     &      & LLava-NeXT & mPLUG-Owl3 & GPT-4o & Qwen2-VL & InstructBLIP \\
\midrule
\multirow{4}*{\makecell[c]{Single-Image\\Recognition}}   
& single-hop  & 82.47 (+7.84) & 81.63 (+9.18) & 90.69 (+4.31) & 82.60 (+8.20) & 68.42 (+36.84) \\
& two-hop     & 71.69 (+9.64) & 90.48 (+57.15) & 89.22 (+5.73) & 80.43 (+26.69) & 62.50 (+30.92) \\
& three-hop   & 77.78 (+18.06) & 89.66 (+62.07) & 86.96 (+5.95) & 71.71 (+18.54) & 50.00 (+16.67) \\
& four-hop    & 71.43 (+11.43) & 75.00 (+53.85) & 86.76 (+7.70) & 72.25 (+23.04) & 50.00 (+24.72) \\
\midrule
Multi-Image Recognition
&         & 78.11 (+1.65) & 77.00 (+16.59) & 93.59 (-0.93) & / & / \\
\midrule
\multirow{2}*{\makecell[c]{Multi-Image\\Retrieval(Forward)}}    
& single-hop  & 11.89 (-1.68) & 80.86 (-0.26) & 83.56 (+1.36) & / & / \\
& two-hop     & 10.06 (-0.31) & 73.82 (+1.89) & 73.51 (+0.86) & / & / \\
\midrule
\multirow{2}*{\makecell[c]{Multi-Image\\Retrieval(Backward)}}  
& single-hop  & 23.87 (+2.74) & 85.29 (-0.14) & 88.71 (+1.53) & / & / \\
& two-hop     & 11.84 (-1.37) & 72.55 (+0.50) & 78.26 (+0.57) & / & / \\
\midrule
\multirow{1}*{\makecell[c]{Knowledge Association}}    
&   & 18.32 (+3.01) & 28.96 (+4.09) & 70.27 (-0.31) & / & / \\
\bottomrule
\end{tabular}
}
\caption{The performance on different reasoning tasks using VE (Visual Consistency Enhancement) prompts. Values in bracelets refer is compared with end-to-end prompts.}
\label{tab:ve}
\end{table*}

\subsubsection{Impact of Task Bias}
In multimodal reasoning tasks, besides the challenges posed by modality conversion and information transmission, task type also plays a key role in performance inconsistency. Specifically, different models may be good at addressing different tasks while neglecting others, leading to notable performance variations across different types of tasks. As can be seen from Table \ref{tab:result}, LLaVA-NeXT achieves high consistency on \textit{Image Recognition} tasks while low consistency on \textit{Image Retrieval} tasks, indicating that it excels at identifying entities in images but performing poorly in retrieving images with given entities, while mPLUG-Owl3 is just the opposite. 

We attribute this phenomena to an imbalance in model training tasks and objectives, where the model fails to comprehensively cover and balance optimization across different types of reasoning tasks (e.g., recognition tasks, retrieval tasks, and compound tasks), causing task-specific inconsistencies in multimodal knowledge reasoning.

\subsubsection{Impact of Reasoning Process}
The reasoning process is an important factor affecting multimodal knowledge reasoning consistency. In this part, we investigate different reasoning processes to assess their performance. We mainly discuss two processes: \textit{Stepwise Prompt in Text} and \textit{Visual Consistency Enhancement Prompt}. The detailed design of these prompts is provided in Appendix \ref{sec:B.3}.

\noindent \textbf{Stepwise Prompt in Text} The Chain-of-Thought (CoT)\citep{wei2022chain, kojima2022large} paradigm, as a step-by-step reasoning approach, has been proven to effectively enhance model performance in complex reasoning tasks. Under purely textual prompts, CoT guides the model to decompose reasoning steps, progressively building a chain of reasoning, thereby reducing the risk of reasoning failure. Therefore, we first introduce a stepwise prompt in text using CoT prompt to enforce our multimodal knowledge reasoning tasks and present the consistency results in Table \ref{tab:stepwise}.

It can be seen that guiding the model to perform CoT during reasoning can improve the consistency to some extent, indicating that a reasonable breakdown of the reasoning chain can help the model complete multimodal knowledge reasoning tasks more consistently. Although end-to-end prompt are more intuitive for humans, they do not show advantages in consistency for MLLMs. Furthermore, as the number of reasoning hops increases, the end-to-end prompt performs even worse and the improvement of stepwise prompt in text becomes even higher, indicating the superiority of CoT in complex multimodal knowledge reasoning.

\noindent \textbf{Visual Consistency Enhancement Prompt} Aside of simple stepwise prompt in text (CoT), we wonder if there is a better reasoning process for MLLMs. Therefore, we investigate visual consistency enhancement prompt. The core idea behind visual enhancement prompting is explicitly decompose the reasoning chain to twp steps, first to extract key visual features through explicit visual recognition and summarization, and then integrate these features into the textual reasoning. 

Specifically, in multimodal knowledge reasoning tasks, this methodology constrains the prompting to explicitly identify all visual inputs and extract key features (such as objects, scenes, or relationships within images) before proceeding with stepwise textual reasoning. This reduces the model's tendency to overly focus on the textual modality or to produce results inconsistent with the visual modality. As demonstrated in Table \ref{tab:ve}, Visual Consistency Enhancement Prompt improves consistency across various tasks, especially on complex reasoning tasks which requires more than one reasoning hop. Models integrating visual enhancement with Chain-of-Thought (CoT) prompting exhibit high consistency across different tasks.

\section{Conclusion}
In this research, we discover the consistency problem in multimodal knowledge reasoning in MLLMs. We construct multiple tasks and design a multi-hop, multi-image, multi-task benchmark for evaluating consistency in multimodal knowledge reasoning. We find that current MLLMs struggle to maintain consistency when faced with complex reasoning task. The analysis further reveals multiple factors affecting consistency, including reasoning hops, relation type, task type and reasoning process, pointing out directions for future research.

\section*{Limitations}
We mainly conduct experiments on five common MLLMs, with more MLLMs unexplored. We only design four multimodal knowledge reasoning tasks, with more complex tasks to be discussed. 

\section*{Ethics Statement}
We use open-source dataset and models as their intended uses and licenses. Our dataset contains photos of celebrities available online with no harmful or private content. We respect everyone's privacy. ChatGPT is used to assist writing only. 

\bibliography{ref}


\appendix


\section{Experiment Details}

\subsection{Multi-image Retrieval and Reasoning Task}
\label{sec:A.1}
To evaluate the consistency of MLLMs in visual and multimodal reasoning tasks, we designed an experimental dual retrieval paradigm, including Forward Retrieval and Backward Retrieval modes, each with three progressive test tasks.

\textbf{Forward Retrieval:}
This stage includes the following three tasks:
\begin{itemize}
    \item \textbf{Visual Retrieval:} Given an image of a target attribute (e.g., the logo of Chicago Bulls) and three distracting images, the model must identify the target attribute ("Which image represents Chicago Bulls?").
    \item \textbf{Text Knowledge Retrieval:} Input a black neutral image , requiring the model to infer the answer based on textual knowledge ("Which team did Michael Jordan play for the longest?").
    \item \textbf{Cross-modal Backward Retrieval:} Reuse the four candidate images from Task 1 and require the model to reverse-locate the visual attribute using entity knowledge ("Which picture represents the team that Michael Jordan played for the longest?").
\end{itemize}

\textbf{Backward Retrieval:}
This stage includes the following three tasks:
\begin{itemize}
    \item \textbf{Visual Retrieval:} Given an image of a target entity (e.g., Michael Jordan) and three distracting images, the model is required to identify the target entity ("Which image shows Michael Jordan?").
    \item \textbf{Text Knowledge Retrieval:} Input a black neutral image and provide four candidate entity names, requiring the model to infer the answer based on textual knowledge ("Which player played for Chicago Bulls?").
    \item \textbf{Cross-modal Forward Retrieval:} Reuse the four candidate images from Task 1 and require the model to combine visual recognition and knowledge reasoning ("Which picture represents the person who played for Chicago Bulls??").
\end{itemize}

In order to increase the difficulty of the task, select pictures of the same type to construct candidate answers. Both tasks are evaluated using the same criterion: if the model can correctly answer Tasks (1) and (2) but fails in Task (3), it indicates an inconsistency between the visual features and semantic knowledge.

\subsection{Cross-modal Knowledge Association Tasks}
\label{sec:A.2}
The Knowledge Association task aims to assess the model's reasoning consistency across multiple cross-modal transformations. This task combines Multi-image Recognition and Backward Retrieval tasks and requires the model to repeatedly perform information association reasoning across multiple modalities. Specifically, it includes the following four sub-tasks:
\begin{itemize}
    \item \textbf{Visual Recognition:} Given several images, the model needs to identify the entity that appears most frequently in these images.
    \item \textbf{Textual Reasoning:} Input a black image and ask, "Which team did Michael Jordan play for the longest time?"
    \item \textbf{Visual Retrieval:} Provide an image of Chicago Bulls and three distracting images, and ask the model to recognize which image represents Chicago Bulls.
    \item \textbf{Cross-modal Reasoning:} Given the images from Task (1) and the images from Task (3), ask the model, "Which of the last four images represents the team that the player who appears most frequently in the first four images played for the longest time?"
\end{itemize}
The key aspect of this task is whether the model can maintain consistency across successive cross-modal reasoning steps. If the model performs well in Tasks (1)-(3) but fails in Task (4), it indicates that there are still limitations in the model's consistency in multiple cross-modal reasoning tasks. The uniqueness of the Knowledge Association task lies in simulating real-world complex reasoning demands, where the model needs to switch between modalities repeatedly and maintain reasoning consistency. This design not only reveals the model's performance in individual tasks but also evaluates its stability in complex reasoning chains.

\begin{table*}[t]
\centering
\resizebox{\textwidth}{!}{
\begin{tabular}{lcccc}
\toprule
\textbf{Model} & \textbf{Visual inconsistency Num} & \textbf{Non-Visual inconsistency Num} & \textbf{Total inconsistency Num} & \textbf{Visual inconsistency Rate (\%)} \\
\midrule
LLava-NeXT & 92 & 317 & 409 & 22.49\% \\
mPLUG-Owl3 & 138 & 646 & 784 & 17.60\% \\
Qwen2-VL & 59 & 243 & 302 & 20.14\% \\
InstructBLIP & 71 & 405 & 476 & 14.92\% \\
GPT-4o & 88 & 243 & 331 & 26.59\% \\
\bottomrule
\end{tabular}
}
\caption{Comparison of inconsistency in single-hop reasoning tasks}
\label{tab:reasoning_tasks}
\end{table*}

\subsection{Alias Matching}
\label{sec:app-alias}
In multimodal reasoning tasks, the model's output may semantically align with the standard answer but differ in vocabulary. Therefore, exact word-level matching is insufficient for accurate assessment. To address this, we extracted synonyms and aliases for each candidate answer from \text{Wikipedia} and created a \text{key-value (KV) table} that includes the candidate answers and their corresponding aliases. Each entry in this table records a candidate answer and its list of synonyms or aliases. Most words in our dataset, such as names of people and places, have clear aliases or variants, effectively covering the diverse expressions the model may use.

During the evaluation process, we compare the model's output with each entry in the \text{candidate answer and alias table}. If the model's output matches any of the candidate answers or their synonyms/aliases, it is considered correct. This approach evaluates the model's ability to reason in natural language based on semantics rather than exact word matching.

\begin{table*}[htbp]
\centering
\begin{tabular}{lp{10cm}}
\toprule
\textbf{Type} & \textbf{Context} \\
\midrule
End-to-End Prompting1 & Give you a picture <image>, please answer the following question, which team did the person in the picture play for the longest time? \\
\midrule
End-to-End Prompting2 & Give you a picture <image>, please complete the following fill-in-the-blank question, the team of the person in the picture played for the longest timeis \_\_\_\_\_\_ \\
\midrule
Stepwise Prompt1 & Give you a picture <image>. Please think carefully and answer the following questions step by step. Which team did the person in the picture play for the longest time? Please give your answer. \\
\midrule
Stepwise Prompt2 & Give you a picture <image>. Please think carefully and answer the following questions. Which team did the person in the picture play for the longest time? Please give your answer step by step. \\
\midrule
Stepwise Prompt3 & You are shown a picture of a person.  <image>. \newline Based on your knowledge of this person, please provide the name of the team that the person played for the longest time. Let's think step by step. \\
\midrule
\multirow{1}*{\makecell[c]{Visual Consistency\\Enhancement Prompt}}& Give you a picture <image> and answer the following questions . \newline
Step 1: Carefully identify who is in the picture. \newline
Step 2: Based on your knowledge of this person, Which team did he play for the longest time? 
Let's think step by step. \\
\bottomrule
\end{tabular}
\caption{Comparison of Different Prompting Methods}
\label{tab:prompting_methods}
\end{table*}

\section{Detailed Experimental Data}

\subsection{Relation Type - Error Rate Comparison Data}
\label{sec:B.1}

\textbf{Model Consistency Analysis in Single-Hop Reasoning Tasks}

We compared the consistency of all models in single-hop reasoning tasks and analyzed the inconsistency distribution across different relation categories, as shown in the figure. We categorized relations such as "sport," "country of citizenship," "position played on team/speciality," "capital," and "religion or worldview" as relations that can be directly inferred from visual information. These relations usually do not require complex background knowledge for inference. On the other hand, relations like "author," "spouse," etc., cannot typically be inferred from visual information and rely on language understanding and knowledge reasoning abilities.

To better visualize the sources of errors in different models, we calculated the proportion of errors related to visual reasoning relations in all errors, as shown in the table \ref{tab:reasoning_tasks}.

\textbf{Results Analysis:}
\begin{itemize}
    \item mPLUG-Owl3 produced the most errors (784 in total), with 82.40\% of them being non-visual errors (646 errors), indicating that its consistency in language understanding tasks is poor.
    \item GPT-4o produced fewer errors overall (331 in total), with the lowest number of non-visual errors (243 errors), but the highest proportion of visual errors (26.59\%), indicating that its consistency in visual reasoning tasks requires improvement.
    \item LLaVA-NeXT and Qwen2-VL performed at an intermediate level, with inconsistencies present in both visual and non-visual tasks, but without the extreme characteristics observed in mPLUG-Owl3 or GPT-4o.
    \item Instruct exhibited high consistency in visual reasoning tasks (the lowest proportion of visual errors, 14.92\%), but had a relatively high total error count (476 errors), with 85.08\% of the errors (405 errors) being non-visual, indicating that its primary source of inconsistency lies in non-visual tasks.
\end{itemize}

These results suggest significant differences in model consistency across visual and non-visual reasoning tasks, further revealing the limitations of current multimodal models in their reasoning capabilities.

\subsection{Prompt Design Templates}
\label{sec:B.3}

In this study, we aim to investigate how different prompt designs affect the consistency of multimodal models in cross-modal reasoning tasks. We hypothesize that the way prompts are phrased can lead to reasoning path breaks, which can cause reasoning inconsistencies. Therefore, we designed a series of experiments to compare how different types of prompt structures influence model consistency. Using the recognition task as an example, we manually constructed, generated with GPT, and selected several types of prompts, as shown in Table \ref{tab:prompting_methods}

We used the single-hop recognition task as the core testing scenario and constructed the following three main prompt formats:
\begin{itemize}
    \item End-to-End Prompting: Directly ask the question in natural language and require the model to complete the full cross-modal reasoning process from visual recognition to textual reasoning in a single inference step.
    \item Stepwise Prompt: Build on the original end-to-end prompt by guiding the model to generate a chain of thought (CoT) during textual reasoning to enhance reasoning stability.
    \item Visual Consistency Enhancement Prompt: Explicitly identify all visual inputs in the prompt, and then perform textual reasoning step by step.
\end{itemize}

\section{Case Study}

\begin{table*}[htbp]
\centering
\resizebox{\textwidth}{!}{
\begin{tabular}{|m{4cm}|m{3cm}|m{3cm}|}
  \hline
  \multirow{3}{*}{\includegraphics[width=3cm]{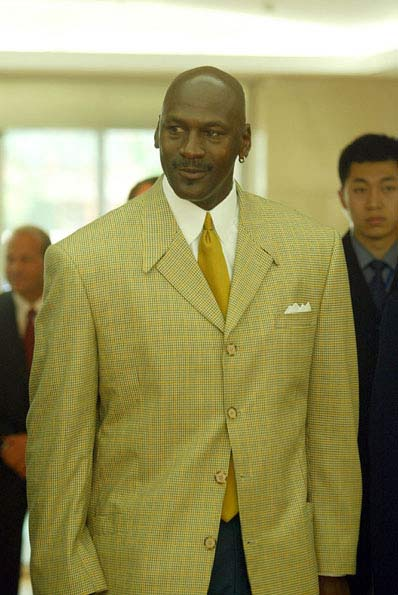}} & Which team did Michael Jordan play for the longest time? & Michael Jordan played for the Chicago Bulls for the longest time \\
  \cline{2-3}
  & Who is the person in the image? & The person in the image is Michael Jordan. \\
  \cline{2-3}
  & What team did the person in the image play for the longest time? & The person in the image played for the Los Angeles Lakers for the longest time \\
  \hline
\end{tabular}
}
\end{table*}

\begin{table*}[htbp]
\centering
\resizebox{\textwidth}{!}{
\begin{tabular}{|m{4cm}|m{3cm}|m{3cm}|}
  \hline
  \multirow{3}{*}{\includegraphics[width=3cm]{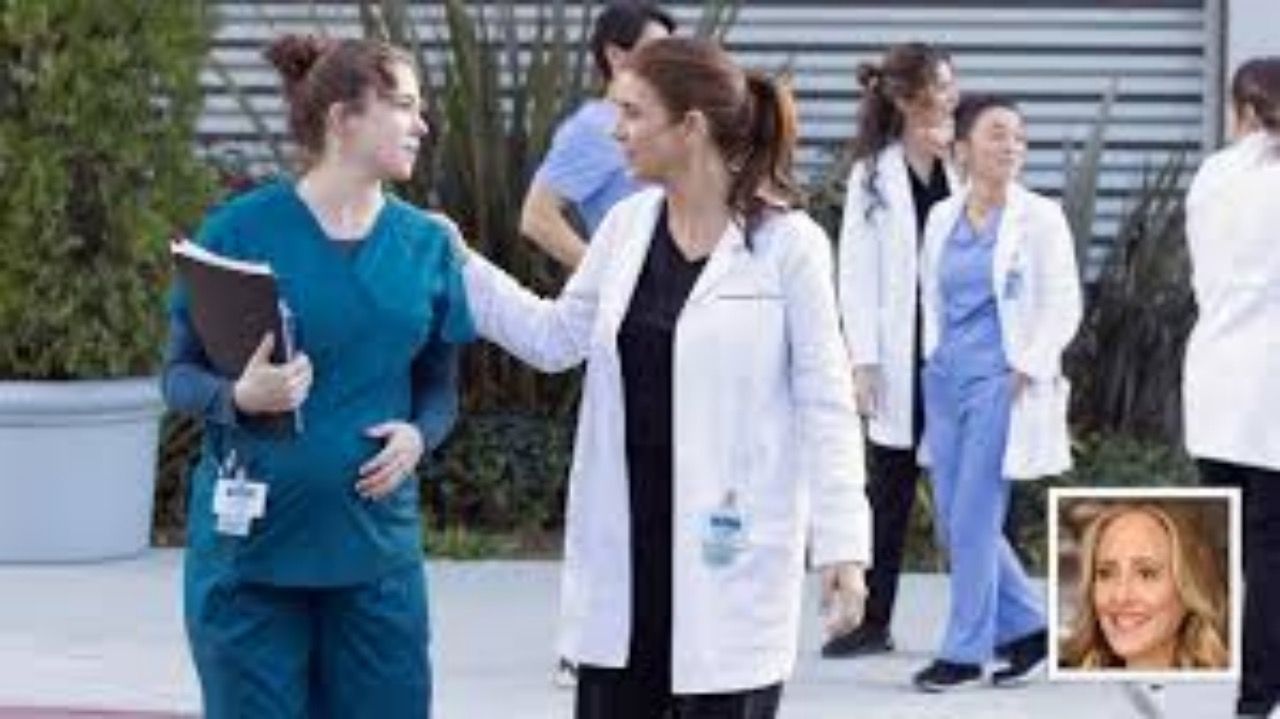}} & Who is the original broadcaster of Grey's Anatomy? & The original broadcaster of Grey's Anatomy is ABC (American Broadcasting Company). 
 \\
  \cline{2-3}
  & What is the TV show in the picture? 
 & The TV show in the picture is "Grey's Anatomy." 
 \\
  \cline{2-3}
  & Who is the original broadcaster of the TV show in the picture?
 &  The image you've provided appears to be from a TV show, but it's not clear which show it is. The characters are dressed in medical scrubs, suggesting they are nurses or doctors, but without more context or a clearer image of the show's title or logo, it's not possible to identify the original broadcaster of the TV show.
 \\
  \hline
\end{tabular}
}
\end{table*}



\end{document}